\documentclass{llncs}
\usepackage{makeidx}  
\usepackage{amssymb}
\usepackage{bbm}

\usepackage[english]{babel}
\usepackage[T1]{fontenc}
\usepackage[utf8]{inputenc}
\usepackage{graphicx, cite}
\usepackage{amsmath}
\graphicspath{{pics/}{figs/}}

\begin{document}
\frontmatter          
\pagestyle{empty}  
\mainmatter              
\title{Retinal Fluid Segmentation and Detection in Optical Coherence Tomography Images using Fully Convolutional Neural Network}
\titlerunning{Short Title}  
\author{Donghuan Lu$^1$, Morgan Heisler$^1$, Sieun Lee$^1$, Gavin Ding$^1$, Marinko V. Sarunic$^1$,and Mirza Faisal Beg$^1$}
\institute{$^1$School of Engineering Science, Simon Fraser University, Burnaby V5A 1S6, Canada}

\maketitle              

\begin{abstract}
As a non-invasive imaging modality, optical coherence tomography (OCT) can provide micrometer-resolution 3D images of retinal structures. Therefore it is commonly used in the diagnosis of retinal diseases associated with edema in and under the retinal layers. In this paper, a new framework is proposed for the task of fluid segmentation and detection in retinal OCT images. Based on the raw images and layers segmented by a graph-cut algorithm, a fully convolutional neural network was trained to recognize and label the fluid pixels. Random forest classification was performed on the segmented fluid regions to detect and reject the falsely labeled fluid regions.  The leave-one-out cross validation experiments on the RETOUCH database show that our method performs well in both segmentation (mean Dice: 0.7317) and detection (mean AUC: 0.985) tasks.
\end{abstract}

\section{Introduction}
\label{sec:intro}
Optical coherence tomography is an established imaging modality in ophthalmology that provides micrometer resolution 3D images of sub-surface biological tissue. It can be used to monitor disease progression, such as in diabetic retinopathy or age-related macular degeneration (AMD). High quality visualizations of the retinal structures provided by OCT images can improve the understanding of the onset and development of these retinal diseases, which are increasing in prevalence and are major causes of visual morbidity \cite{joussen2010retinal}. As such, tools to evaluate the health of the retina non-invasively and quantitatively are urgently needed. Previously, our group used machine learning to segment the retinal vasculature in OCT angiograms \cite{prentavsic2016segmentation} with promising results. Accurate segmentations of regions of cystic macular edema (denoted as "fluid") are also of interest to ophthalmic clinicians. Although 3D OCT images can visualize these regions of fluid, quantitative measurements of size in response to treatment require automated computational algorithms.

Fully convolutional neural networks (FCN) have demonstrated excellent performance for image segmentation tasks\cite{long2015fully}. An advanced version, the U-net\cite{ronneberger2015u}, has proven to out-perform other methods in the application of segmenting small data sets of medical images, such as neural structures\cite{ronneberger2015u}, the kidney\cite{cciccek20163d} and liver tumors\cite{christ2017automatic}. In this paper, we present a novel FCN-based framework for the segmentation and detection of retinal fluid in OCT images. The network used for segmentation followed the U-net structure detailed in Section \ref{sec:layers}, but took additional spatial information as input. Segmentation results were further improved by random forest classifiers trained on the candidate fluid regions.  

\section{Methods}
\label{sec:method}
Our framework for the segmentation and classification of retinal fluid consisted of three steps: 1) Layer segmentation - pre-process the image and segment the internal limiting membrane(ILM) and the retinal pigment epithelium (RPE); 2) Fluid detection - detect potential fluid regions using fully convolutional neural network(FCN); 3) Classification - extract features from potential fluid regions and train a classifier to reject false fluid regions.
\subsection{Materials}
\label{ssec:data}
\begin{figure*}[tbh!]
\centering
\begin{center}
\begin{tabular}{ccc}
\includegraphics[width=4cm, height=2.5cm]{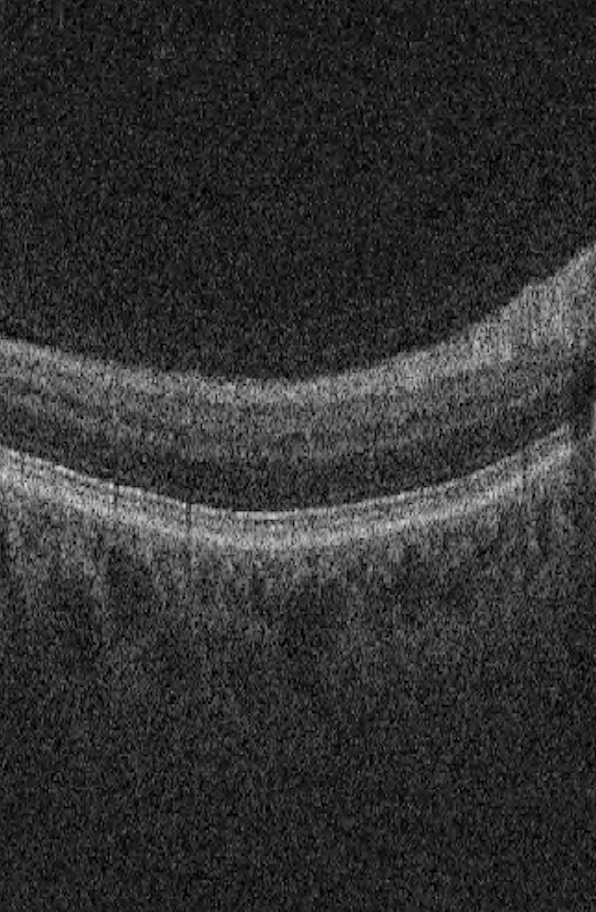} & \includegraphics[width=4cm, height=2.5cm]{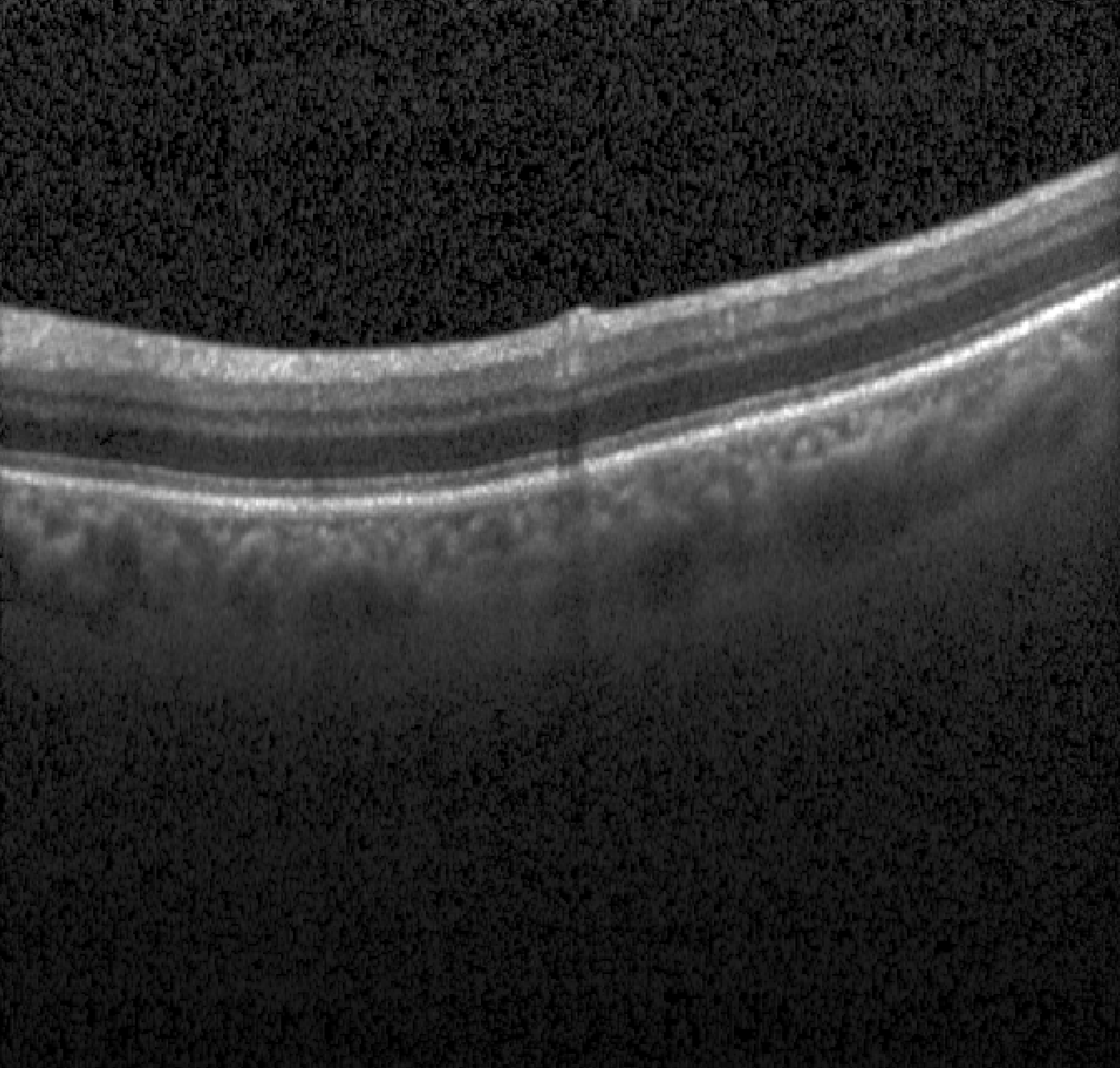} & \includegraphics[width=4cm, height=2.5cm]{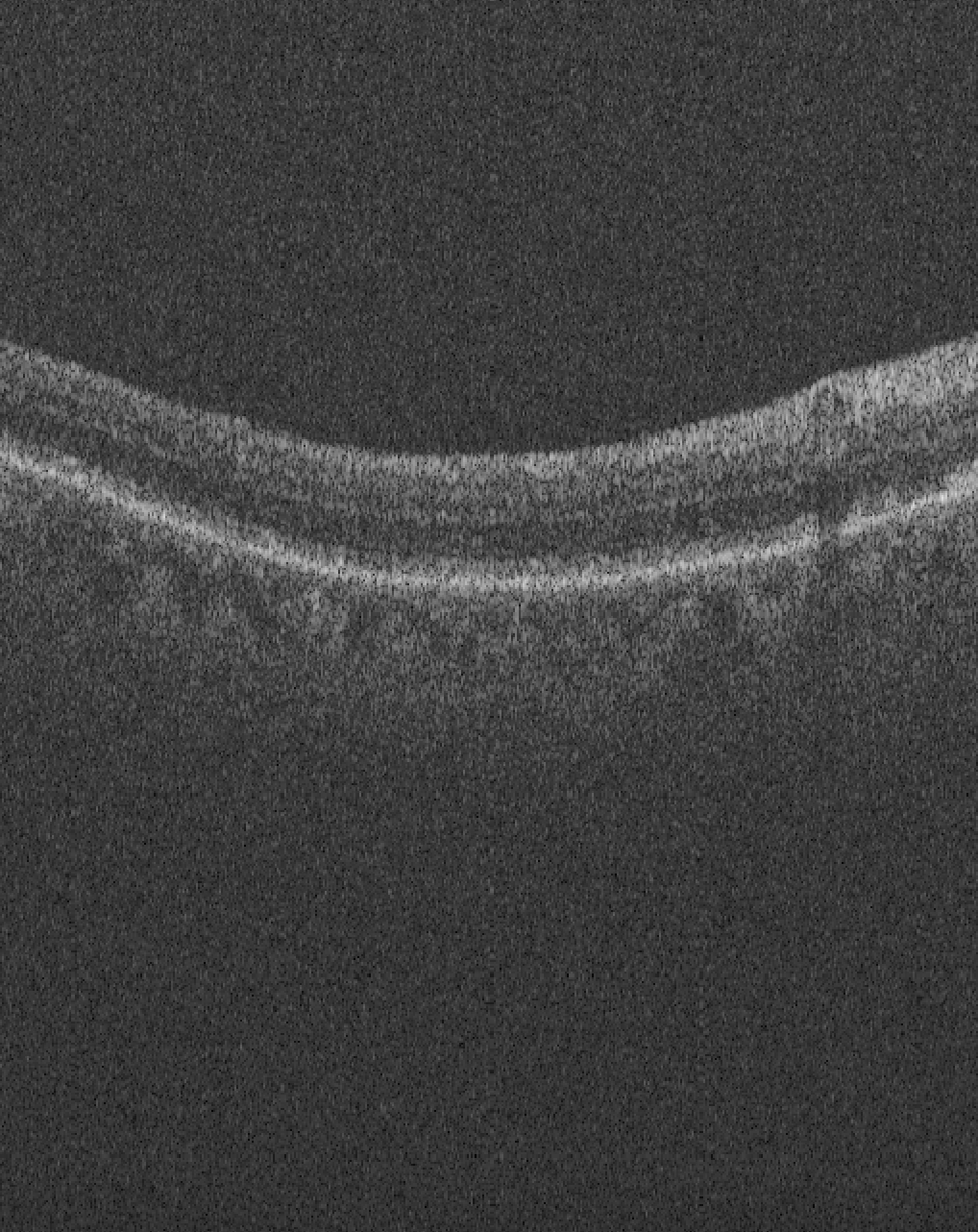}\\
\includegraphics[width=4cm, height=2.5cm]{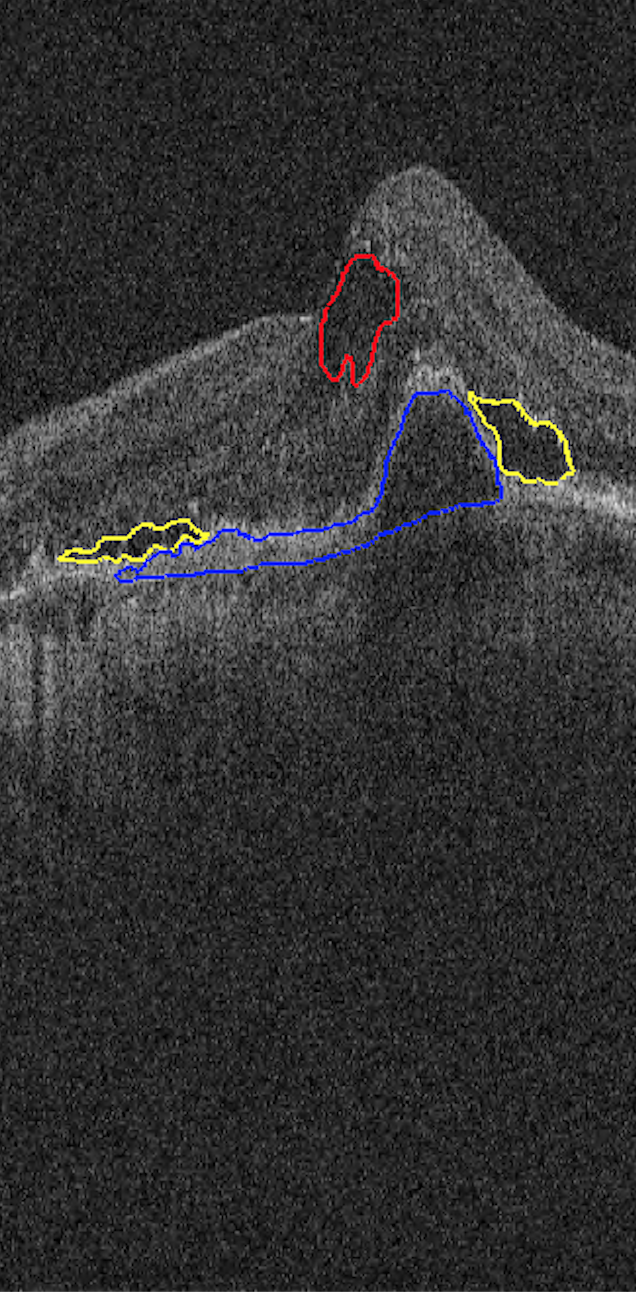} & \includegraphics[width=4cm, height=2.5cm]{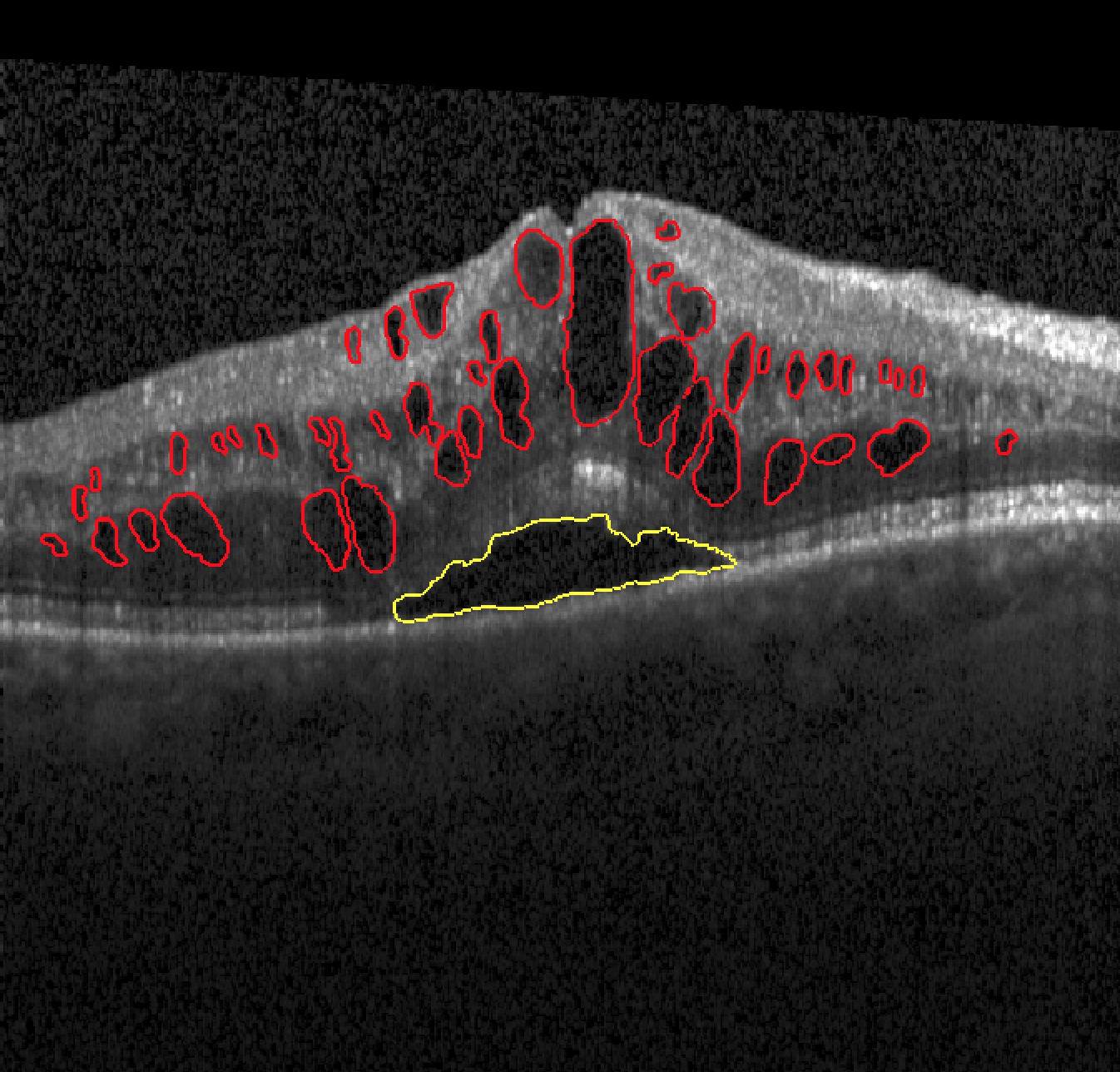} & \includegraphics[width=4cm, height=2.5cm]{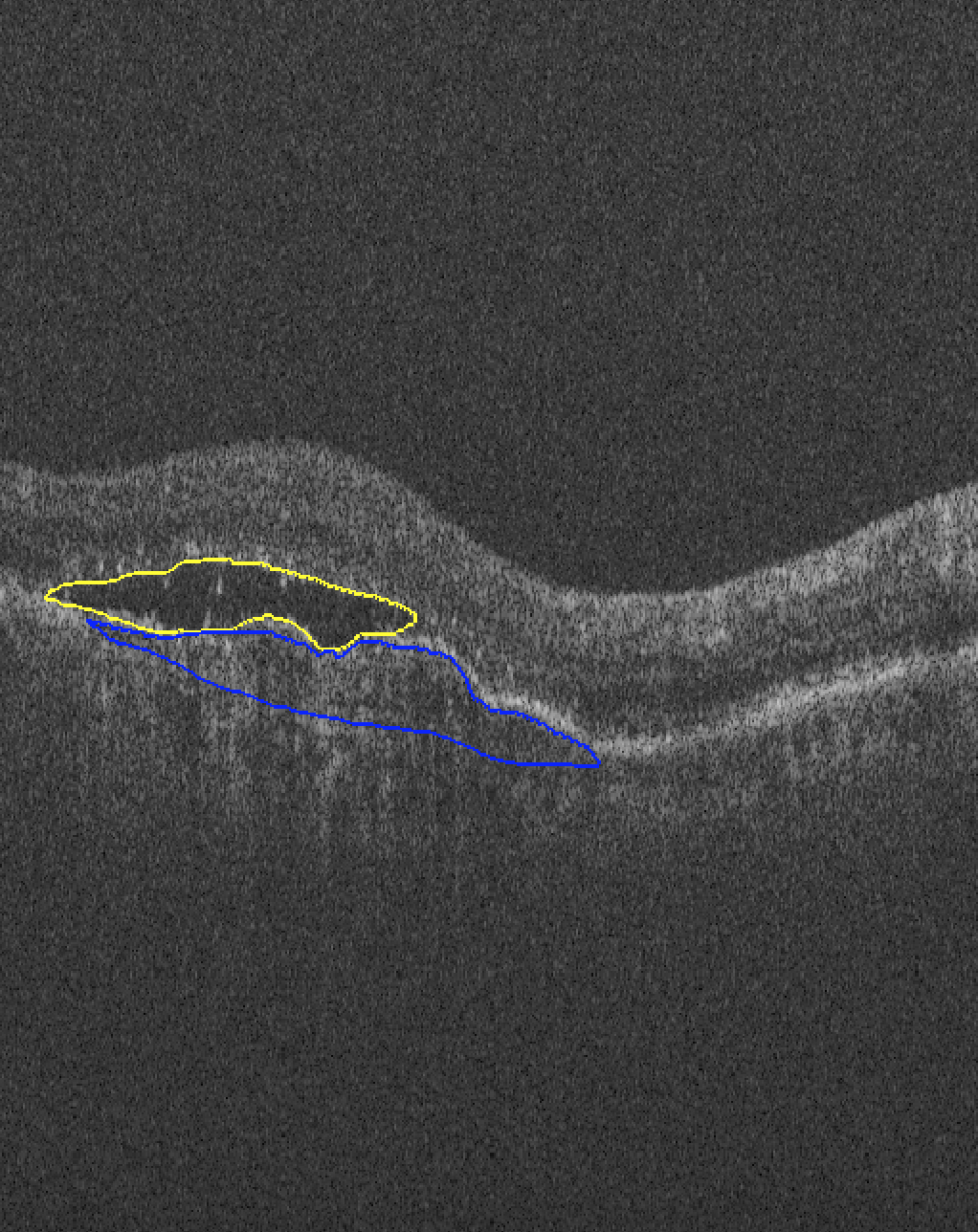}\\
(a) Cirrus & (b) Spectralis & (c) Topcon
\end{tabular}
\end{center}
\caption{OCT images with manual segmentation. The images on the first row have no macular edema, while the red, blue and yellow segmentations on the second row represent the IRF, PED and SRF respectively.}
\label{fig:Example}
\end{figure*}

The images used in this study are provided by the MICCAI RETOUCH Group\cite{WinNT}. There were 3 training data sets and a total of 70 volumes, with 24 volumes acquired with each of the two OCT imaging devices: Cirrus (Zeiss) and Spectralis (Heidelberg), and 22 volumes acquired with T-1000 and T-2000 (Topcon, collectively referred to by the manufacturer name in the rest of paper). For each volume from these three devices, the numbers of B-scans were 128, 49 and 128, respectively. Three different types of fluid, namely the intraretinal fluid (IRF), subretinal fluid (SRF) and the pigment epithelial detachment (PED) were manually labeled and provided as ground truth. Although not all B-scans contained fluid, there was at least one type of fluid in each volume. Topcon images with and without macular edema are shown in Fig \ref{fig:Example} for each commercial device.

\subsection{Layer Segmentation}
\label{sec:layers}
Axial motion between scans was corrected using cross-correlation. Bounded variation (BV) 3D smoothing was applied to the motion-corrected intensity B-scans in order to reduce the effect of speckle while preserving and enhancing the boundaries between retinal layers. Next, the ILM and RPE were automatically segmented using a 3D graph-cut based algorithm \cite{doi:10.1167/iovs.12-11521}.

\subsection{Fully Convolutional Neural Network}
\begin{figure}
\vspace{0.1cm}
\centerline{\includegraphics[width=12cm]{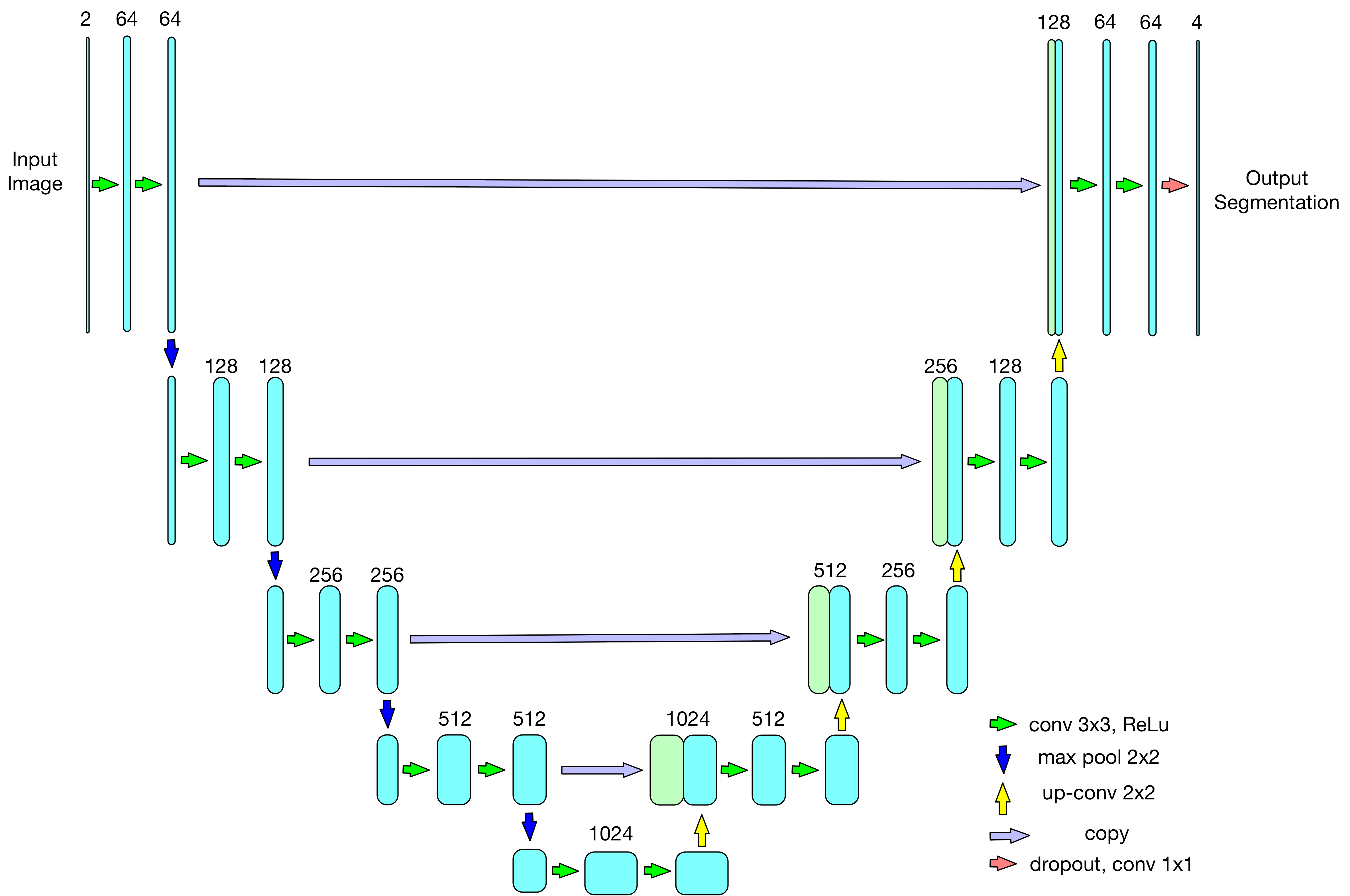}}
\caption{Fully convolutional neural network. Each number above the cyan box represents the number of channels of the feature map.}
\label{fig:network}
\end{figure}

The network architecture is illustrated in Fig \ref{fig:network}. It shared a similar structure as the standard U-net\cite{ronneberger2015u} except the input image contained a second channel in addition to the raw image. Relative distance maps were concatenated to the raw image as the second channel based on the assumption that the location of the fluid within the retina was an important property to determine the type of fluid. Since a network that classifies each pixel solely by the intensities of its neighbours cannot capture this information, additional inputs were necessary to better classify the different fluid types. For a pixel $(x,y)$ in the relative distance map, its intensity in the relative distance map is defined as:
\begin{equation} \label{eq:location}
I(x,y)=\frac{y-Y_{1}(x)}{Y_{1}(x)-Y_{2}(x)} \quad,
\end{equation}
where $Y_{1}(x)$ and $Y_{2}(x)$ represent the $y$-coordinate of ILM and RPE, respectively.

There were two paths in the network architecture: the contracting path (left side) and the expansive path (right side). Each path consisted of 4 blocks. In each block there were two convolutional layers with kernel size $3\times 3$ and a rectified linear unit (ReLU) after each convolution operation. A $2\times 2$ max pooling layer with stride 2 was then added to the contracting path and a $2\times 2$ up-convolution layer was added to the expansive path. Shortcut connections were added to the layers with the same resolution from the contracting path to the expansive path to provide high-resolution features. After the expansive path, a $1\times 1$ convolutional layer was used to map the features to a 4 channel probability map corresponding to background, IRF, SRF and PED. For each pixel, the channel with the highest probability was chosen as the segmentation result.

The network was trained end-to-end with a pixel-wise softmax function $p_j(\vec{z})={e^{z_j}}/{\sum_{k=1}^{4}e^{z_k}}$ combined with a cross entropy loss function:
\begin{equation} \label{eq:lossfunction}
H(i)=-\frac{1}{N}\sum_{i=1}^{N}\sum_{j=1}^{4}[\mathbbm{1}\{y^i=j\}log(h(x^i)_j]\quad,
\end{equation}
where $z_j$ denotes the probability in channel $j$, $N$ is the number of input samples, $j$ represents the class of samples, $x^i,y^i$ are the feature vector and label of the $i_{th}$ sample and $h$ represents the network function.

To avoid overfitting, a dropout layer\cite{srivastava2014dropout} was inserted before the $1\times 1$ convolutional layer. During the training stage, only half of the units were randomly retained to feed features to the next layer, while in the testing stage, all the units were kept to generate the segmentation. Without training all units on every sample, it reduced overfitting by preventing co-adaption on the training data. Because there were far more background pixels than fluid pixels, the training set was highly imbalanced. Training with equation \ref{eq:lossfunction} would result in a network tending to predict most pixels as background. Therefore, the network was trained only with true positive and false positive pixels with a learning rate of $10^{-4}$.  

One important advantage of FCN is that without any fully connected layer, the network can be applied to images of arbitrary size. In this study, although the sizes of images from 3 devices were not the same, the network trained on one data set was used to initialize the networks for the other two data sets to accelerate the training process.Due to the limited number of training samples, data augmentation is essential to prevent overfitting and increase the robustness and invariance properties of the network. Three processes - flip, rotation and zooming - were applied to the training samples. The rotation degree was from $-25^\circ$ to $25^\circ$ and the maximum zooming ratio was $0.5$.
\subsection{Random Forest Classification}
Because the network tended to over-segment, random forest classifiers\cite{breiman2001random} were trained to rule out false positive regions and determine the presence of fluid for each volume\cite{lu2017multiple}. In every B-scan of the given volume, potential fluid pixels with 8-connectivity were defined as a candidate region, and regions with less than 3 pixels were removed. For each candidate region, a bounding rectangle whose edge was 1.2 times the tight bounding box was extracted from which a 16 dimension feature vector was then further extracted. The features included the major and minor axis lengths, ratio of major and minor axis lengths, perimeter, area, ratio of perimeter and area, eccentricity, orientation, variance of the cyst height at each column, average intensity inside the cyst, average intensity outside the cyst, intensity difference of inside and outside, intensity variance inside, intensity kurtosis inside, intensity skewness inside, and the relative distance of the center pixel as defined in equation \ref{eq:location}. The label of candidate region was defined by $r=\frac{area(S_1 \cap S_2)}{min(area(S_1),area(S_2))}$, where $S_1$ is the segmented region and $S_2$ is the manual segmentation. The candidate region was labeled as truth when $r>0.7$, or vice versa.

Three random forest classifiers were trained for each type of fluid separately. To avoid the effect of imbalanced training samples, the weight of positive samples and negative samples were set as $N(negative)/(N(negative)+N(positive))$ and $N(positive)/(N(negative)+N(positive))$ respectively, where $N(\cdot)$ represents the number of samples. The output probability of the random forest classifier was compared with a threshold to determine the label of each sample. A threshold of 0 to 1 with an interval of 0.01 was tested by a 5-fold cross validation on the training set and the one which resulted in the maximum $Fmeasure=\frac{2\times precision \times recall}{precision+recall}$ was selected.

To detect the presence of fluid in volume $k$, first the probability of each B-scan containing fluid was defined as the highest probability of all candidate regions within the scan. As fluid usually existed within multiple B-scans, the mean of the 10 highest probabilities over all B-scans in the volume were calculated to determine the presence of fluid.

\section{Experiments and Results}
The deep neural network was built with an open source deep learning toolbox, Tensorflow\cite{tensorflow2015-whitepaper}. To evaluate the performance of our proposed approach, leave-one-out experiments were performed on each data set. At every iteration, a single volume was used for testing and the rest were used to train the neural network and 3 random forest classifiers corresponding to each type of fluid. The segmentation performance was evaluated by Dice index and absolute volume difference(AVD):
\begin{equation}
\begin{aligned}
&Dice=\frac{2area(S_1 \cap S_2)}{area(S_1)+area(S_2)} \\
&AVD=|area(S_1)-area(S_2)|
\end{aligned}
\end{equation}

\begin{figure*}[tbh!]
\centering
\begin{center}
\begin{tabular}{ccc}
\includegraphics[width=4cm,height=3.5cm]{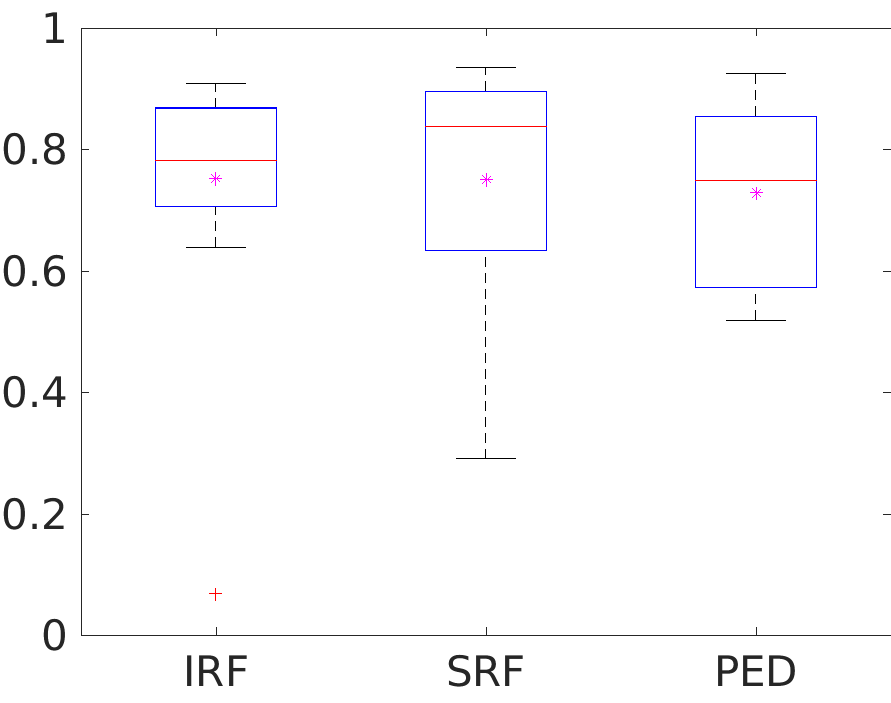} &
\includegraphics[width=4cm,height=3.5cm]{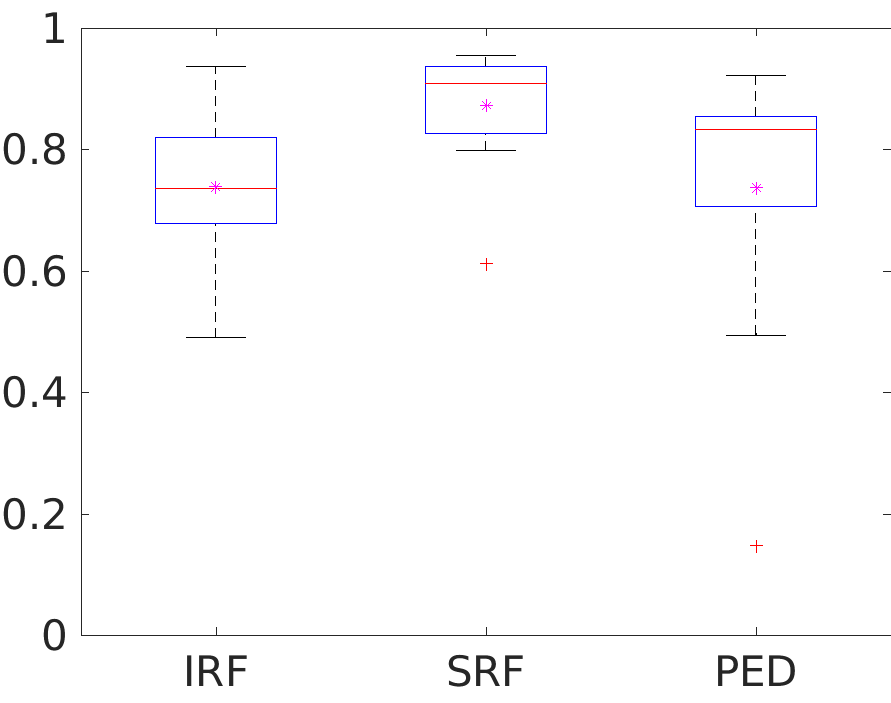} &
\includegraphics[width=4cm,height=3.5cm]{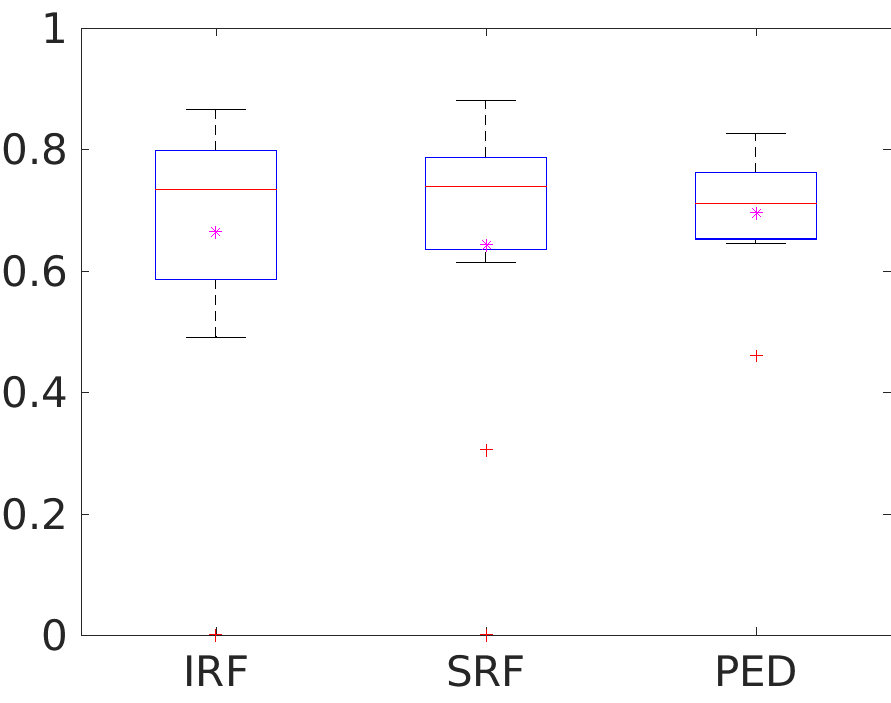}
\\
(a) Cirrus & (b) Spectralis & (c) Topcon
\end{tabular}
\end{center}
\caption{Boxplot of the Dice index for detected fluid regions in each of the three commercial devices. The stars in each box are the mean of Dice and the line represents the median Dice index.}
\label{fig:Dice}
\end{figure*}

\begin{figure*}[tbh!]
\centering
\begin{center}
\begin{tabular}{ll}
\includegraphics[width=6cm,height=4cm]{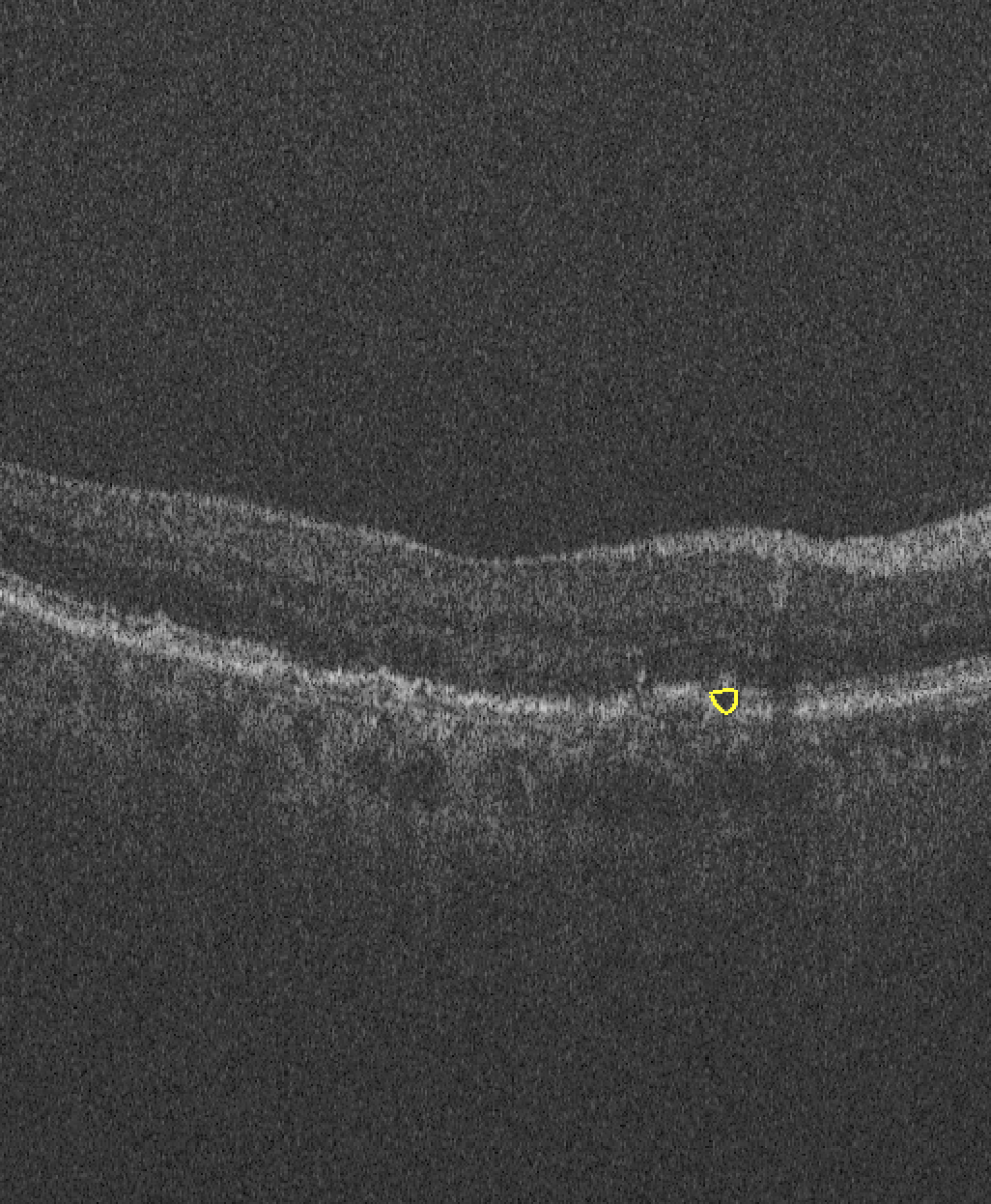} &
\includegraphics[width=6cm,height=4cm]{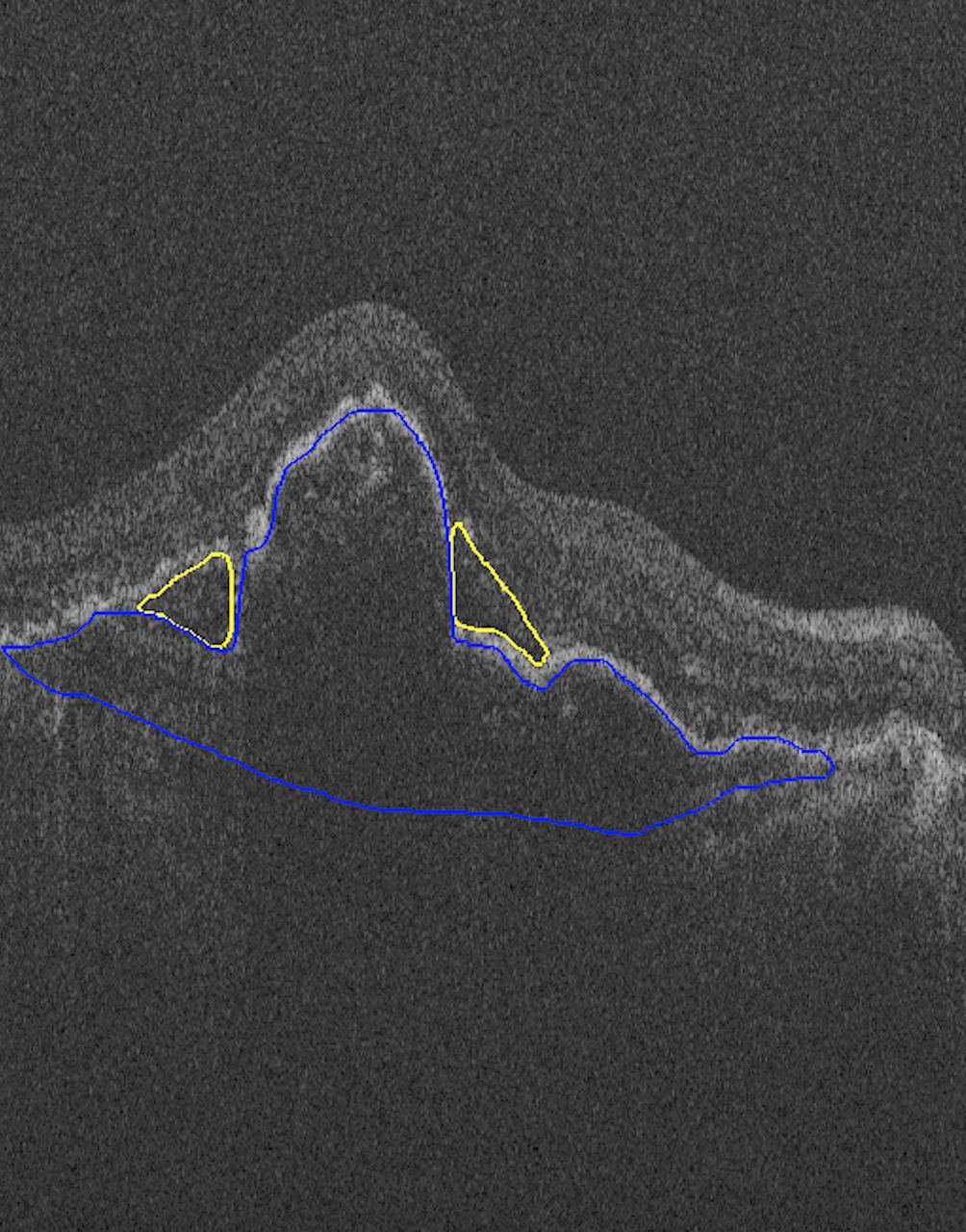}
\end{tabular}
\end{center}
\caption{Illustration of outliers. The yellow and blue lines indicate manual segmentation of SRF and PED. For the left image, the SRF is too small to be detected, while for the right image, some of the PED pixels are very close to the background.}
\label{fig:outlier}
\end{figure*}

Both Dice and AVD were computed per volume. For each type of fluid, the segmentation was measured separately, and volumes which doesn't contain this type of fluid were ignored for its measurement. The mean and standard deviation of AVD is presented in Table \ref{table:avd} and the Dice index was displayed in Fig \ref{fig:Dice}. The red $'+'$ indicates outliers that were caused by two reasons: 1) the volume only contained a very small area of fluid and the proposed method failed to detect it, such as left panel in Fig \ref{fig:outlier}; 2) the retina was warped too much, such as right panel in Fig \ref{fig:outlier} due to extremely severe disease and the proposed method could not recognize all of the fluid pixels, as there was no similar volume with which to train the network in the leave-one-out experiment. 
\begin{table}[h]
\small
\centering
\setlength{\tabcolsep}{0.1in}
\begin{center}
\begin{tabular}{cccc}
\multicolumn{1}{c}{Device} &\multicolumn{1}{c}{\bf IRF} &\multicolumn{1}{c}{\bf  SRF} &\multicolumn{1}{c}{\bf PED}\\ \hline 
Cirrus  &94649(223420)  &142250(390700) &122610(291820)\\
Spectralis  &12759(17888) &5186(5626) &14729(14704)\\
Topcon &25924(28032) &25042(38640) &162520(359970)\\
\hline
\end{tabular}
\end{center}
\caption{Absolute volume difference shown as mean(standard deviation).}
\label{table:avd}
\end{table}

\begin{figure*}[tbh!]
\centering
\begin{center}
\begin{tabular}{ccc}
\includegraphics[width=4cm,height=3.5cm]{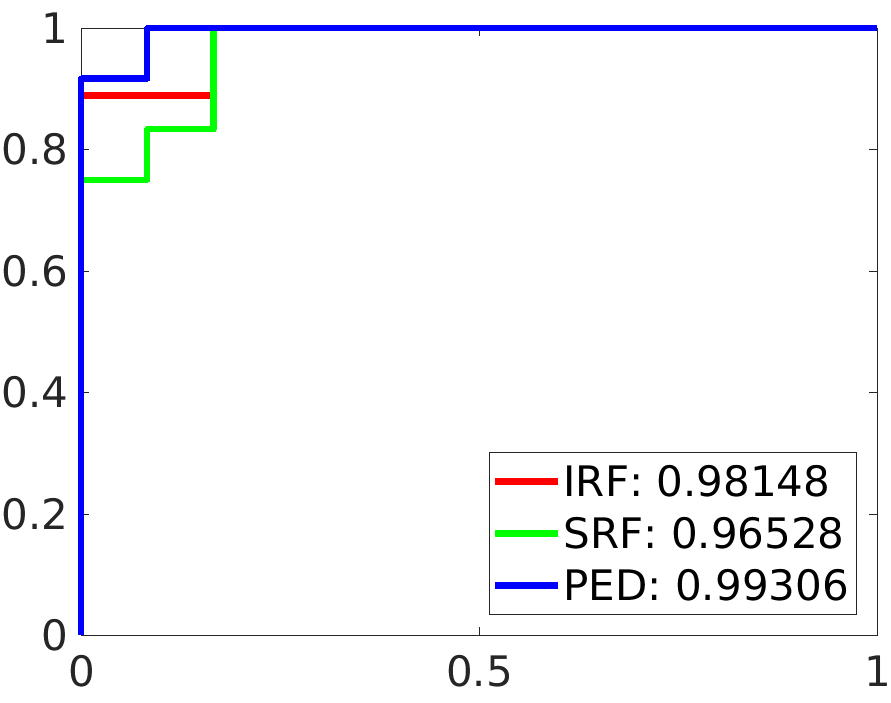} &
\includegraphics[width=4cm,height=3.5cm]{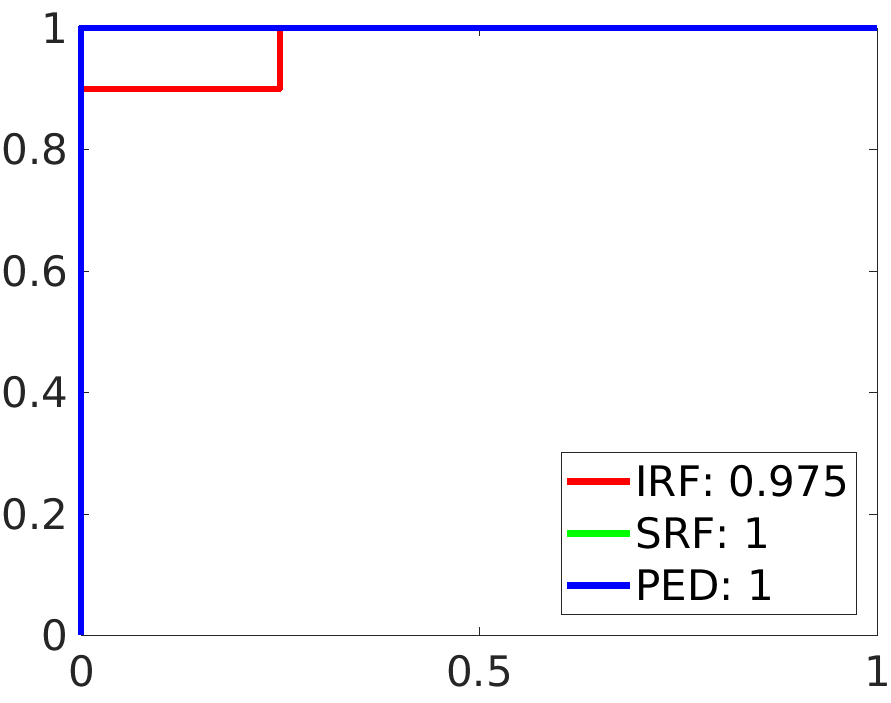} &
\includegraphics[width=4cm,height=3.5cm]{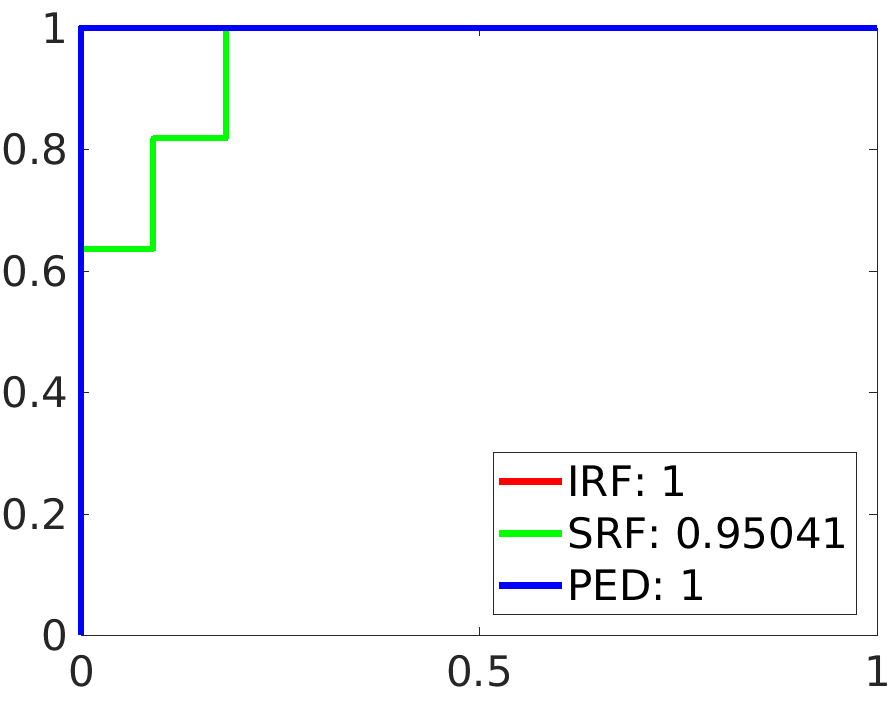}
\\
(a) Cirrus & (b) Spectralis & (c) Topcon
\end{tabular}
\end{center}
\caption{ROC curve for comparing the detected fluid to the manual segmentations. The numbers in legend are the AUC of IRF, SRF and PED.}
\label{fig:ROC}
\end{figure*}

For the detection of fluid present in each volume, the Receiver Operating Characteristic (ROC) curve was created by comparing the probabilities to the manual label and the area under the curve (AUC) was calculated. The probability of each volume was defined as the mean probability of 10 B-scans that mostly likely contained that type of fluid. The result is displayed in Fig \ref{fig:ROC}. 

\section{Conclusion}
\label{sec:conclusion}
In this paper, we described a novel framework to automatically segment the fluid regions in 3D retina OCT images, and detect different fluids types. Based on ILM and RPE layer segmentations using a graph-cut technique, a relative distance map was generated and concatenated with the raw image as an input to train a fully convolutional deep neural network. Pixels classified as potential fluid by the network were then grouped into different regions based on their 8-connectivity. For each type of fluid, a random forest classifier was trained on those candidate regions to rule out false positive samples and determine the fluid presence in each volume. The proposed new method showed good performance in the leave-one-out experiments with 3 different training sets. Due to the limited number of training samples in the given data sets, the segmentation algorithm performed poorly on some volumes. As more data is accumulated in the future, we expect better accuracy in both fluid segmentation and detection.
\section{Acknowledgements}
\label{sec:acknowledgements}
The authors would like to acknowledge funding support from the Natural Sciences and Engineering Research Council of Canada (NSERC), Canadian Institutes for Health Research (CIHR), the Brain Canada Foundation, Alzheimer Society of Canada, the Pacific Alzheimer Research Foundation, Genome British Columbia, and the Michael Smith Foundation for Health Research (MSFHR).

\bibliographystyle{splncs03}
\bibliography{main}

\begin{thebibliography}{10}
\providecommand{\url}[1]{\texttt{#1}}
\providecommand{\urlprefix}{URL }

\bibitem{WinNT}
the retinal oct fluid challenge, \url{http://retouch.grand-challenge.org}

\bibitem{tensorflow2015-whitepaper}
Abadi, M., Agarwal, A., Barham, P., et~al.: {TensorFlow}: Large-scale machine
  learning on heterogeneous systems (2015), \url{http://tensorflow.org/},
  software available from tensorflow.org

\bibitem{breiman2001random}
Breiman, L.: Random forests. Machine learning  45(1),  5--32 (2001)

\bibitem{christ2017automatic}
Christ, P.F., Ettlinger, F., Gr{\"u}n, F., Elshaera, M.E.A., Lipkova, J.,
  Schlecht, S., Ahmaddy, F., Tatavarty, S., Bickel, M., Bilic, P., et~al.:
  Automatic liver and tumor segmentation of ct and mri volumes using cascaded
  fully convolutional neural networks. arXiv preprint arXiv:1702.05970  (2017)

\bibitem{cciccek20163d}
{\c{C}}i{\c{c}}ek, {\"O}., Abdulkadir, A., Lienkamp, S.S., Brox, T.,
  Ronneberger, O.: 3d u-net: learning dense volumetric segmentation from sparse
  annotation. In: International Conference on Medical Image Computing and
  Computer-Assisted Intervention. pp. 424--432. Springer (2016)

\bibitem{joussen2010retinal}
Joussen, A., Gardner, T., Kirchhof, B., Ryan, S.: Retinal Vascular Disease.
  Springer Berlin Heidelberg (2010),
  \url{https://books.google.ca/books?id=rGj7EWTxOP4C}

\bibitem{lang2015automatic}
Lang, A., Carass, A.e.a.: Automatic segmentation of microcystic macular edema
  in oct. Biomedical optics express  6(1),  155--169 (2015)

\bibitem{doi:10.1167/iovs.12-11521}
Lee, S., Fallah, N., Forooghian, F., Ko, A., Pakzad-Vaezi, K., Merkur, A.B.,
  Kirker, A.W., Albiani, D.A., Young, M., Sarunic, M.V., Beg, M.F.: Comparative
  analysis of repeatability of manual and automated choroidal thickness
  measurements in nonneovascular age-related macular degeneration.
  Investigative Ophthalmology \& Visual Science  54(4),  2864 (2013)

\bibitem{long2015fully}
Long, J., Shelhamer, E., Darrell, T.: Fully convolutional networks for semantic
  segmentation. In: Proceedings of the IEEE Conference on Computer Vision and
  Pattern Recognition. pp. 3431--3440 (2015)

\bibitem{lu2017multiple}
Lu, D., Ding, W., Merkur, A., Sarunic, M.V., Beg, M.F.: Multiple instance
  learning for age-related macular degeneration diagnosis in optical coherence
  tomography images. In: Biomedical Imaging (ISBI 2017), 2017 IEEE 14th
  International Symposium on. pp. 139--142. IEEE (2017)

\bibitem{pilch2013automated}
Pilch, M., Stieger, K.e.a.: Automated segmentation of pathological cavities in
  optical coherence tomography scanspathological cavities in oct scans.
  Investigative ophthalmology \& visual science  54(6),  4385--4393 (2013)

\bibitem{prentavsic2016segmentation}
Prenta{\v{s}}i{\'c}, P., Heisler, M., Mammo, Z., Lee, S., Merkur, A., Navajas,
  E., Beg, M.F., {\v{S}}aruni{\'c}, M., Lon{\v{c}}ari{\'c}, S.: Segmentation of
  the foveal microvasculature using deep learning networks. Journal of
  biomedical optics  21(7),  075008--075008 (2016)

\bibitem{ronneberger2015u}
Ronneberger, O., Fischer, P., Brox, T.: U-net: Convolutional networks for
  biomedical image segmentation. In: International Conference on Medical Image
  Computing and Computer-Assisted Intervention. pp. 234--241. Springer (2015)

\bibitem{srivastava2014dropout}
Srivastava, N., Hinton, G.E., Krizhevsky, A., Sutskever, I., Salakhutdinov, R.:
  Dropout: a simple way to prevent neural networks from overfitting. Journal of
  Machine Learning Research  15(1),  1929--1958 (2014)

\bibitem{wang2016automated}
Wang, J., Zhang, M., Pechauer, A.D., Liu, L., Hwang, T.S., Wilson, D.J., Li,
  D., Jia, Y.: Automated volumetric segmentation of retinal fluid on optical
  coherence tomography. Biomedical optics express  7(4),  1577--1589 (2016)

\bibitem{wilkins2012automated}
Wilkins, G.R., Houghton, O.M., Oldenburg, A.L.: Automated segmentation of
  intraretinal cystoid fluid in optical coherence tomography. IEEE Transactions
  on Biomedical Engineering  59(4),  1109--1114 (2012)

\end{thebibliography}

\end{document}